\pdfoutput=1
\documentclass[journal]{IEEEtran}
\usepackage{cite}
\usepackage{amsmath,amssymb,bm}
\usepackage{graphicx}
\usepackage{placeins}
\usepackage{booktabs,tabularx,array}
\usepackage{siunitx}
\usepackage{url}
\newcommand{\vect}[1]{\boldsymbol{#1}}
\newcommand{\matr}[1]{\mathbf{#1}}

\title{Fiber Bragg Grating Whiskers for Bioinspired Hydrodynamic Perception on Underwater Robots}
\author{Hao~Li$^{1,*}$, Tianyu~Tu$^{1}$, Siyue~Yao$^{2}$, Ziyang~Chang$^{3}$, Juhyun~Jung$^{1}$,\\
Xiaochi~Xie$^{1}$, Long~Yin~Chung$^{1}$, Tian-Ao~Ren$^{1}$, Genliang~Chen$^{2}$, and Mark~Cutkosky$^{1,*}$%
\thanks{$^{1}$Stanford University.}%
\thanks{$^{2}$Shanghai Jiao Tong University.}%
\thanks{$^{3}$Tsinghua University.}%
\thanks{$^{*}$Corresponding authors: Hao Li and Mark Cutkosky.}%
\thanks{This work has been submitted to the IEEE for possible publication. Copyright may be transferred without notice, after which this version may no longer be accessible.}}
\begin{document}
\maketitle
\bstctlcite{BSTcontrol}

\begin{abstract}
Harbor seals track hydrodynamic trails with their vibrissae, enabling passive perception of moving targets in dark or turbid water. Inspired by this capability, we present compact fiber Bragg grating (FBG) whiskers for underwater robots. Like seal whiskers, they have a non-uniform taper and elliptical cross-section. Controlled towing experiments show a monotonic relative-flow response from 0.1 to 0.6~m/s, a strong reduction of self-induced oscillation relative to a cylindrical baseline, and a pronounced dependence on angle of attack. Experiments with a pitching foil show that the whiskers can detect the characteristic vortices shed by a stationary or moving source, detectable several seconds after the source has passed.
Using this information, a single front-mounted whisker enabled
a small underwater robot to distinguish between continuing straight
and executing a turn, selecting the correct branch in 17 of 20 trials (85.0\%)
from whisker signals alone.
These results connect bioinspired hydrodynamic sensing to robot action and suggest the utility of whiskers for tracking underwater objects.
\end{abstract}
\section{Introduction}

\noindent
Underwater robots often operate where conventional sensors are degraded. Light attenuates with depth and turbidity, suspended particles cause backscatter, specular surfaces confound photometric pipelines, and acoustic sensing can be power-intensive, geometry-ambiguous, or undesirable in ecologically sensitive settings~\cite{schrope2002whale}. Marine mammals provide a complementary model for perception in such environments: harbor seals (\textit{Phoca vitulina}) can use their whiskers both to detect surfaces as they brush against them and to follow hydrodynamic trails left by moving objects, even when vision is blocked and after delays of ten seconds or more between wake generation and sensing~\cite{dehnhardt2001science,wieskotten2010movingdir}.
These capabilities motivate a robotic analogue in which compact, passive whisker sensors provide local flow context and wake cues when vision is unreliable and active sensing is limited.
A key biophysical insight is morphological. Seal vibrissae are not just thin shafts; they have an elliptical cross-section, and they taper with an undulating pattern such that streamwise phase shifts suppress vortex-induced vibrations (VIV) during swimming~\cite{hanke2010viv}. By reducing self-induced oscillations, this morphology improves the signal-to-noise ratio for detecting external hydrodynamic disturbances. As noted in the next section, whisker analogues have been explored for flow-speed estimation, directional response, oscillatory-flow sensing, and wake detection using various transduction technologies. The most relevant are summarized in Table~\ref{tab:whisker-comparison}.
Despite this progress, translating whisker sensing into robot behavior remains challenging. A robot-mounted whisker must be rugged, compact, and compatible with onboard data acquisition; its signals must remain interpretable under vehicle motion; and the sensing pipeline must produce cues that can drive action in real time. Equally important, the equivalent task is not generic unsteady flow detection, but trail-following after the passage of a moving source.
Here we present an FBG-based whisker sensing system designed for an underwater robot. Controlled towing experiments establish relative flow-speed response, vortex-induced-vibration suppression, and angle-of-attack dependence. Experiments with a pitching foil show that the wake regime of a moving object creates a different wake pattern than a stationary flapper and produces a signature trail that can inform a robot whether to continue straight or to turn when tracking it.
\begin{figure*}[!t]
	\centering
    \includegraphics[width=\textwidth]{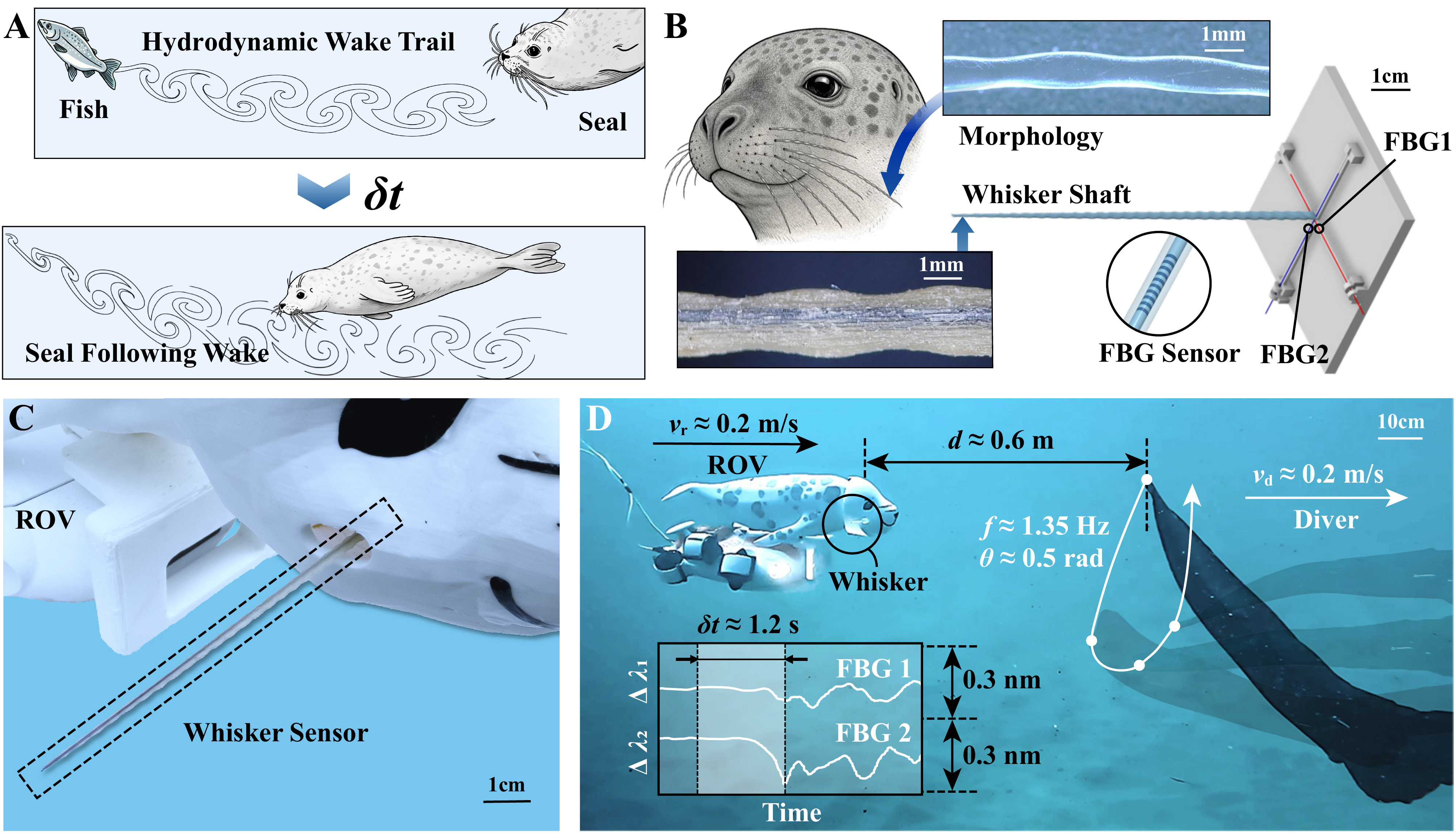}
	\caption{\textbf{
    Bioinspired whisker design and integration onto a robot in the field.}
		(\textbf{A}) Hydrodynamic trail following: vortices shed by a swimming fish persist in water and may be encountered by a trailing seal after $\delta t$. This delay depends on the relative position and motion of source and observer, as well as on how long coherent vortices remain detectable.
        (\textbf{B}) Harbor seal whisker morphology and the biomimetic whisker module. (Top) Seal whiskers exhibit an elliptical cross-section and undulating taper from base to tip [adapted from Murphy et al.~\cite{murphy2013angle}, CC BY 4.0.]  (Bottom-left) Detail of whisker shaft in this work.
        (Right) Whisker sensor based on Fiber Bragg Gratings (FBGs). Whisker shaft and bridge structure are designed to transfer bending moments to FBGs embedded in the base.
        (\textbf{C}) Detail of a whisker mounted to the nose-piece of a ROV.
        (\textbf{D}) Ocean field deployment. The ROV (carrying whisker sensors on both sides of a plastic seal-model housing) follows a human diver while recording FBG wavelength shifts $\Delta\lambda_1$ and $\Delta\lambda_2$ (lower-left trace). $\delta t$ denotes the elapsed time between a fin stroke and the arrival of the corresponding wake-induced spike at the whisker. The diver's fin oscillates with frequency $f$ and angular amplitude $\theta$ (measured from the neutral position to peak excursion). $v_\text{d}$ and $v_\text{r}$ denote the swimming speeds of the diver and the ROV, respectively, and $d$ is the standoff distance between the diver's fin and the whisker sensor.}
	\label{fig:overall}
\end{figure*}

\noindent\textbf{Contributions.} We present the following contributions.
\begin{itemize} \setlength{\itemsep}{0pt}
\item A robust whisker technology suitable for marine robots: a robot-mountable whisker with two-axis fiber Bragg grating (FBG) strain readout, designed such that multiple whiskers can be connected to a single optical fiber. The whiskers employ a superelastic NiTi core and a urethane shell so that they recover after being bent flat (as in a collision) without noticeable change in signal.
\item Validated flow-sensing: calibrated relative flow-speed response, angle-of-attack dependence, and reduction in vortex-induced vibration compared with a cylindrical baseline. In a field demonstration in the ocean, a small remotely operated underwater
vehicle (ROV) was able to detect the vortices shed by the fin of a human diver.
\item A task-relevant wake study: demonstration of the ability to characterize the different characteristic wake patterns created by stationary and moving pitching foil. Building on this capability a small ROV was able to detect the vortices shed by a moving, pitching foil to determine whether it should proceed straight ahead or turn.
\end{itemize}
\subsection{Related Work}
\subsubsection{Biology of hydrodynamic trail following and VIV suppression}
Biological studies established that seals can follow artificial and biogenic wakes using their vibrissae, even when vision is blocked and after delays between wake generation and sensing~\cite{dehnhardt2001science,wieskotten2010movingdir} (Fig.~\ref{fig:overall}A). Subsequent work connected this capability to vibrissal morphology: undulating, elliptical cross-sections with streamwise phase shifts (Fig.~\ref{fig:overall}B) reduce coherent vortex shedding and suppress vortex-induced vibrations during forward swimming~\cite{hanke2010viv}. Reviews synthesize these mechanisms and motivate biomimetic implementations for engineering sensing and tracking~\cite{zheng2021creating}.
\subsubsection{Whisker-inspired flow sensors}
Robotic analogues span capacitive, piezoresistive, piezoelectric, triboelectric, magnetic, optical, and fiber-optic transduction. Many systems demonstrate steady-flow estimation and directional response~\cite{stocking2010capacitance,eberhardt2011bio,y2012design,alvarado2013performance,y2014whisker,kottapalli2014harbor,kottapalli2015harbor,zhang2021harbor}, and some report wake sensing using upstream cylinders or fin-like paddles~\cite{beem2012calibration,beem2015wake,liu2023artificial,zheng2021optimizing,zheng20223d,gul2018fully,wang2021bionic,wang2022underwater,dai2024biomimetic,elshalakani2020deep,glick2024tracking,wang2024bioinspired,xu2024deep,liu2025deep}. These studies establish important sensing abilities,
but in general do not show the ability to detect or track a trail of vortices, as seals do. Table~\ref{tab:whisker-comparison} provides a representative comparison of prior systems most relevant to underwater sensing and robot deployment.
\subsubsection{FBG-based whiskers and underwater robotic deployment}
FBGs have several advantages as a transduction technology for marine applications. First, optical fibers are electrically passive, compatible with saltwater environments, and unaffected by electromagnetic interference. For similar reasons, FBGs have been used in harsh-environment sensing, including downhole oil-industry monitoring~\cite{nellen2003reliability,qiao2017fiber}.
Second, they are highly sensitive: typical strain responses are on the order of \(1~\mathrm{pm}/\mu\varepsilon\), and microstrain-scale strain resolution has been demonstrated~\cite{chen2011review,cusano2004dynamic}.
Third, multiple FBGs with distinct nominal wavelengths can be multiplexed along a single optical fiber and sampled at kHz rates; the optical interrogator can also be placed remotely from the sensing location. Prior FBG-based artificial whiskers have demonstrated underwater hydrodynamic sensing, including wake sensing and direction-related signatures~\cite{glick2024tracking,wang2024bioinspired}. Our work builds on these foundations with a robot-mountable design in which a single fiber can collect signals from multiple whiskers, using two FBGs mounted on orthogonal bridge structures at the base of each whisker to resolve two-axis bending (Fig.~\ref{fig:overall}B). In this paper, we deploy whiskers singly or in pairs on a small ROV with controlled tests in a pool and in a field test in coastal water (Fig.~\ref{fig:overall}, C and D).

\section{Whisker Design and Fabrication}
\begin{figure*}[!t]
	\centering
	\includegraphics[width=\textwidth]{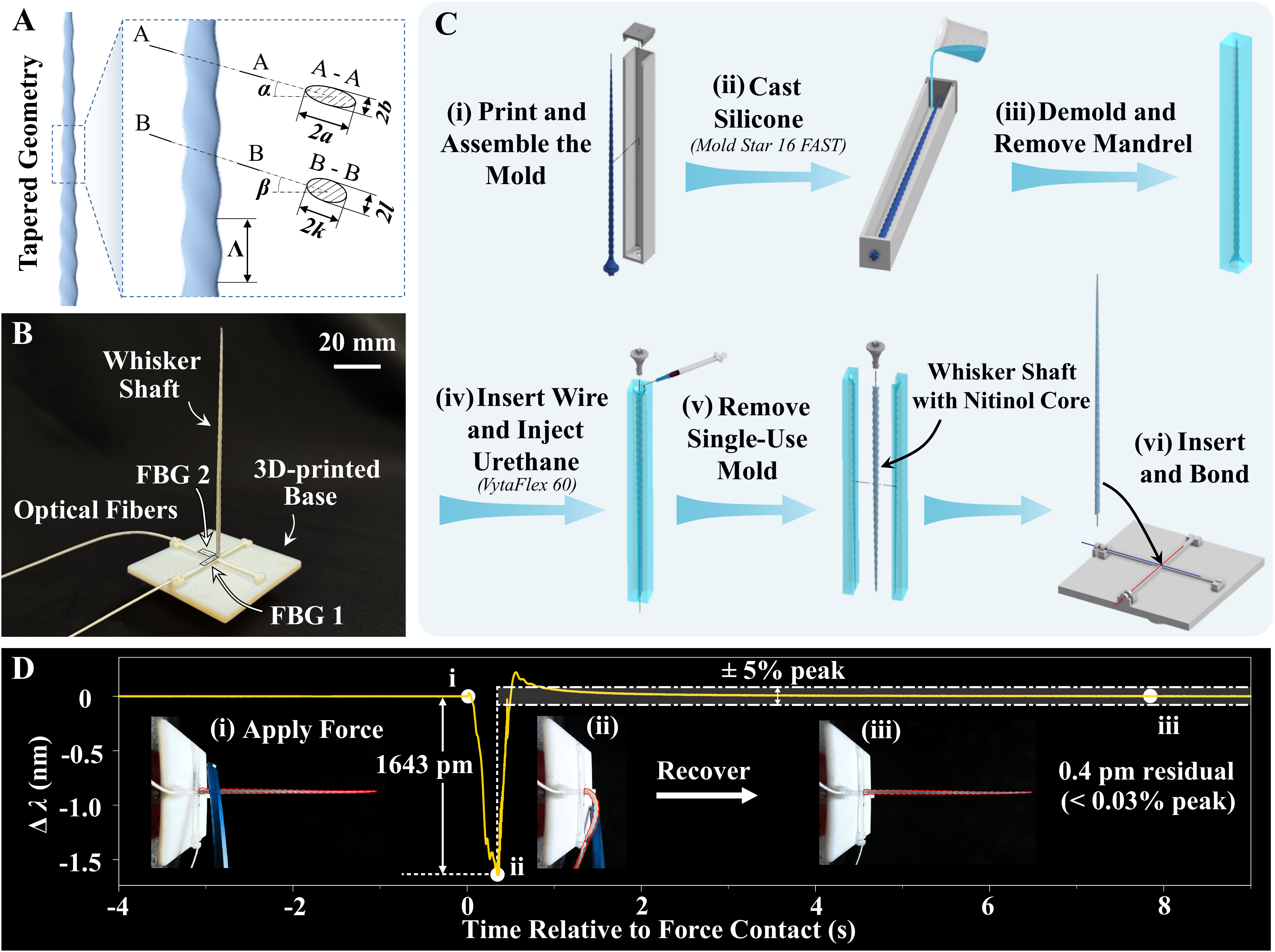}
    \caption{
    \textbf{Whisker morphology, fabrication process, and characterization.}
        (\textbf{A}) Following \cite{hanke2010viv}, the whisker profile is described by the semi-axis lengths of the peak ($a$, $b$) and trough ($k$, $l$) elliptical cross-sections, the inclination angles $\alpha$ and $\beta$, and the undulation wavelength $\Lambda$.
        (\textbf{B}) Photograph of an assembled whisker sensor. One or two optical fibers carrying FBG~1 and FBG~2 are seated in orthogonal grooves on a cross-bridge structure atop a 3D-printed base. The NiTi core of the whisker shaft is bonded at two points: near the cross-bridge center and at the base surface.
        (\textbf{C}) Molding workflow used to fabricate the whisker shaft.
        (\textbf{D}) Recovery characterization under large underwater deflection. Three video frames are overlaid on the FBG~1 wavelength trace, with the whisker shaft outlined in red for visibility.
		}
	\label{fig:whisker_design_fab}
\end{figure*}
When a bluff body is towed through water at a steady velocity, vortex shedding can occur in its wake, which is known as a Kármán vortex street, inducing vortex-induced vibrations (VIV) on the body itself. As shown in Video~1, VIV is clearly observable when a cylindrical whisker is towed through water.
Biological studies have characterized the morphology of the harbor seal
vibrissae in detail and demonstrated that their undulating,
elliptical cross-section suppresses VIV during forward
swimming~\cite{hanke2010viv}.
To take advantage of this geometry, we adopted the parametric model of
Hanke \emph{et al.}~\cite{hanke2010viv}, defining four elliptical
semi-axes ($a$, $b$, $k$, $l$), two inclination angles
($\alpha$, $\beta$), and an undulation wavelength $\Lambda$.
The cross-section has semi-axes with aspect ratios $a/b = 2.95$ at the peak and $k/l = 1.90$ at the trough, producing an alternating pattern of elongated and rounder elliptical profiles
(Fig.~\ref{fig:whisker_design_fab}A).
The whiskers manufactured for the experiments reported here have cross-sectional dimensions approximately three times
those reported in \cite{rinehart2017characterization} and the same inclination angles, $\alpha$ and $\beta$, as reported in \cite{hanke2010viv}.
Further observations suggest that the seal whisker's taper, which increases gradually from base to tip, also improves hydrodynamic performance~\cite{kamat2024undulating}.
Our own whiskers approximate this effect with a quadratic taper.
The total whisker length is about $100$~mm, consistent with the range reported for harbor seal vibrissae~\cite{graff2024three, zheng2025wonders}.
These parameters were implemented in a parametric CAD model
to guide fabrication. Full geometric parameters are provided in Table~\ref{tab:whisker-morphology-params}.
Figure~\ref{fig:whisker_design_fab}B summarizes the underwater whisker sensor and its main components.
Each whisker sensor consists of a whisker shaft and a 3D-printed base that houses two FBGs on orthogonal cross-bridge structures. The shaft comprises a superelastic ASTM F2063 NiTi wire (Ø0.4~mm) overmolded with a urethane elastomer (VytaFlex~60) to create the undulating profile. Unlike monolithic 3D-printed whiskers \cite{wang2024bioinspired, liu2023artificial, kamat2024undulating, zheng2021optimizing, geng2025biomimetic, eberhardt2016development}, this bi-material design combines high flexibility with long-term durability, enabling full recovery from large deformations typical of collisions. When deflected past $90^\circ$ underwater and released, the whisker returns to within $\pm 5\%$ of its equilibrium position (relative to peak deflection) in
under 1~s (Fig.~\ref{fig:whisker_design_fab}D).
We fabricated the whiskers using a two-stage molding workflow (Fig.~\ref{fig:whisker_design_fab}C; Video~8). Tooling and structural parts were
produced on an Objet24 (Stratasys) using VeroWhitePlus (nominal accuracy $\approx$ 0.1~mm). We first 3D-printed a mandrel with the target geometry and placed it inside a printed mold box. After applying a spray-on mold release (Ease Release 200, Mann), we cast silicone rubber (Mold Star 16 FAST) to form a
single-use compliant mold. A nitinol (NiTi) wire was inserted into the silicone mold, and the urethane overmold (VytaFlex 60) was cast and cured. Finally, the silicone mold was peeled away to release the finished whisker.
The fabricated whisker was inserted and bonded with cyanoacrylate adhesive to a printed base that transfers whisker bending into strains on an embedded optical fiber. The base consists of two orthogonal bridge structures with top grooves that locate and secure a single-mode fiber (Corning SMF-28; 0.125~mm cladding diameter, 0.29~mm coated diameter). The NiTi core was bonded at two points: near the center of the cross-bridge and to the underlying base surface, so that bending moments at the whisker root are transferred to both FBGs (Fig.~\ref{fig:whisker_design_fab}B). FBGs (T\&S Communications) were positioned along the fiber adjacent to the whisker anchor locations. Bending moments at the whisker base deform the printed structure, generating tensile or compressive bending strains in the corresponding FBGs. Two gratings resolve bending moments about two mutually perpendicular in-plane axes at the whisker base. With this design, a single fiber with multiple FBGs can be routed through the bridge structures of several whiskers.
\begin{figure*}[!t]
	\centering
	\includegraphics[width=\textwidth]{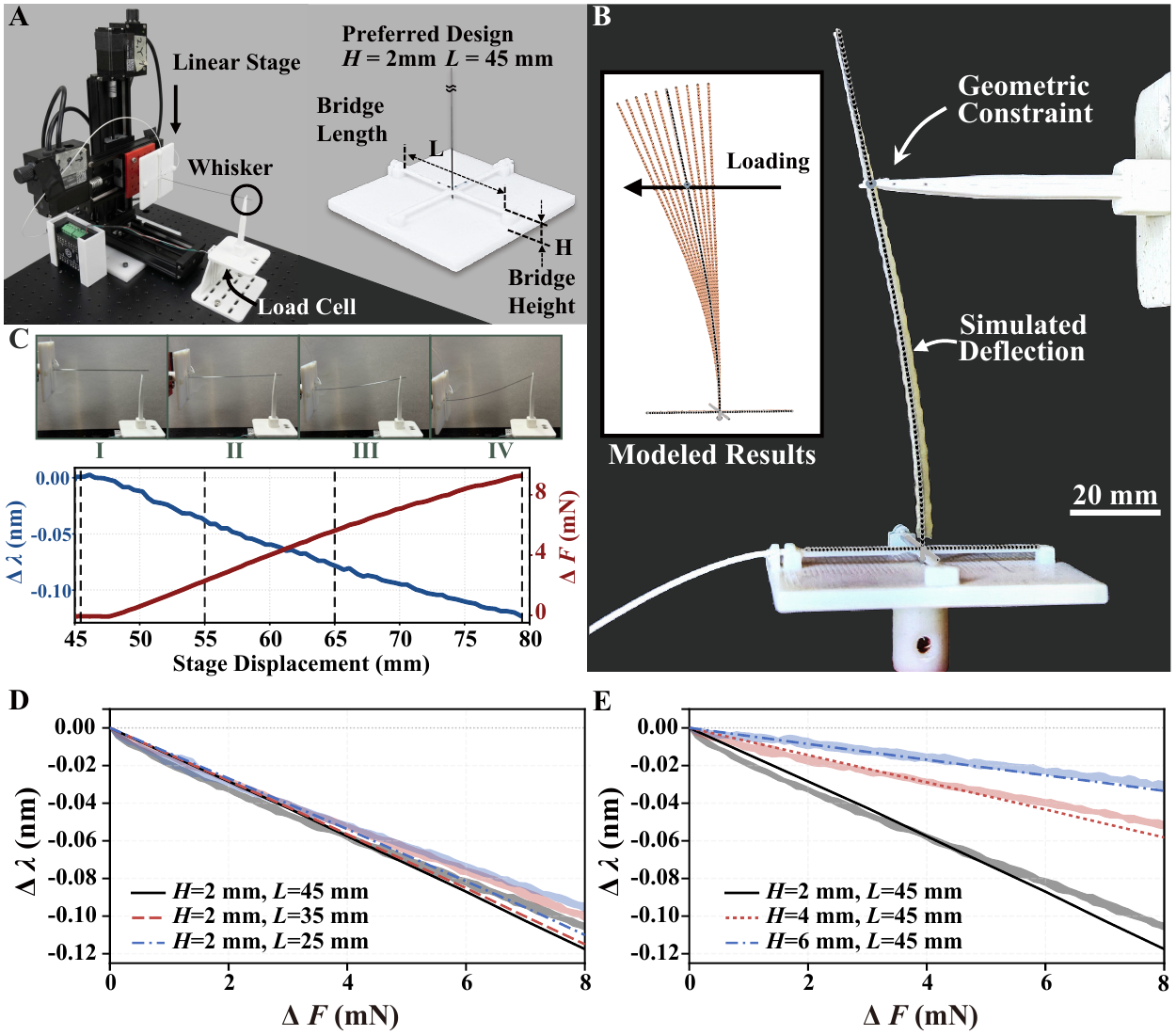}
    
	\caption{\textbf{
    Parametric design for the whisker.
    }
		(\textbf{A}) The testing setup uses a linear stage to apply motions and a load cell to acquire the contact force magnitude.
        (\textbf{B}) A typical comparison between the captured photo and the simulated result of the whisker under a specific load. The inset shows simulated results under different loads.
       (\textbf{C})
       Successive stages of a deflected whisker and plots of contact forces and wavelength shifts. Dotted vertical lines denote states (I-IV) corresponding to photos.
        (\textbf{D, E}) Sensitivity curves of designs with different bridge lengths and heights.
        }
	\label{fig:whisker_model}
\end{figure*}
\section{Analytical Model and Parametric Design}
To tune the sensitivity of the whisker reacting to incoming vortices, we developed a parametric numerical model.
Our approach began with an idealization of the loads: when a vortex propagates through water and reaches the whisker, it induces a time-varying distributed load that we approximated as a time-varying force at some location along the whisker. Because direct underwater calibration is challenging, we performed calibration in air with controlled load location and magnitude.
Figure \ref{fig:whisker_model}A depicts the experimental setup where the whisker is mounted on a vertical slider, with a Y-shaped fork attached to a load cell to provide a known contact location and force.
As the slider moves downward, the contact alters, changing the loading conditions and whisker profile. We recorded the FBG wavelength ($\lambda_{i}$), the load magnitude ($F$), and the slider position ($d$) throughout the process, from before contact until the whisker tip had been deflected by more than 30\,mm. Deflections of the whisker under these conditions are shown in Fig.~\ref{fig:whisker_model}B.
Figure \ref{fig:whisker_model}C illustrates the variation of $\Delta \lambda_{1}$ and $F$ with respect to $d$ during the whole process.
The whisker is composed of multiple materials, forming a flexible system with varying stiffnesses across a large range.
In particular, the height and length of the base bridge, denoted as $H$ and $L$, respectively, are design parameters that influence the sensing performance. Although a multimaterial finite element analysis could provide accurate predictions of deformation and strain, we desire a faster model for rapid design iteration. Accordingly, we adopted a discretized elastic model, presented in \cite{Chen2018Modeling}
in which flexible elements are decomposed into short sections, each of which is modeled as an equivalent 6-DoF elastic mechanism.
The complete whisker-plus-base structure was represented as a hybrid parallel-serial elastic mechanism; implementation details are summarized in Appendix~\ref{app:model-classifier}.
Ultimately, the deformation is transformed into a set of algebraic equations:
\begin{equation}\label{equ::kineto_whisker_equs}
    \vect{c} \left( \bm{\Psi} \right) =
    \matr{K}_{s} \cdot \bm{\Psi} -
    \sum_{i=1}^{n_{0}} \matr{J}_{i}^{T} \cdot \vect{F}_{i} -
    \sum_{i=1}^{4} \matr{J}_{c}^{T} \cdot\vect{F}_{B,i} = \vect{0}
\end{equation}
where
$n_{0}$ is the total number of segments.
$\bm{\Psi} \in \mathbb{R}^{6 \cdot n_{0} \times 1}$ represents joint variables.
Each joint variable contains six components correlated to the six DoFs of the equivalent mechanism of the segment.
$\matr{J}_{i} \in \mathbb{R}^{6 \times 6 \cdot n_{0}}$ ($i=1,2,...,n_{0}$) is the Jacobian matrix mapping from the mechanism joint space to the workspace.
$\matr{J}_{c}$ is the Jacobian matrix relating to the connected point.
$\matr{K}_{s} = \mathrm{diag} (\matr{K}_{1},\matr{K}_{2},..,\matr{K}_{n_{0}}) \in \mathbb{R}^{6 \cdot n_{0} \times 6 \cdot n_{0}}$ denotes the overall structural stiffness matrix.
$\matr{K}_{i}$ ($i=1,2,...,n_{0}$) is the diagonal stiffness matrix of the equivalent segments.
$\vect{F}_{B,i}$ ($i=1,2,3,4$) represents the interaction forces and torques between the base bridge and the whisker, which are correlated with the intrinsic force-deflection behavior.
This relationship can be further expressed as
\begin{equation}\label{equ::kineto_bridge_equs}
    \vect{c}_{i} \left( \bm{\Psi}_{B,i}, \vect{F}_{B,i} \right) = \left[ \renewcommand\arraystretch{1.0}
    \begin{array}{c}
        \left( \mathrm{log} \left(
        \matr{g}_{st} (\bm{\Psi}_{B,i}) \cdot
        \matr{g}_{st} \left( \bm{\Psi} \right)^{-1} \right) \right)^{\vee} \\
        \matr{K}_{s,B,i} \cdot \bm{\Psi}_{B,i} - \matr{J}_{t}^{T} \cdot \vect{F}_{B,i} \\
    \end{array}\right] = \vect{0}
\end{equation}
where $i \in \{1,2,3,4\}$.
Superscript Vee denotes the vector space representation of the log map of the Lie Group.
$\bm{\Psi}_{B,i} \in \mathbb{R}^{6 \cdot n_{i} \times 1}$ represents joint variables,
$n_{i}$ is the total number of segments of the bridge,
$\matr{J}_{t}$ is the Jacobian matrix at the tip end, and
$\matr{K}_{s,B,i}$ is the structural stiffness matrix of the bridge.
Figure \ref{fig:whisker_model}B shows a comparison between the calculated (solid black line) and the actual (gray plot) deflections of one of the whisker design schemes ($H=$ 2\,mm, $L=$ 45\,mm) under loading.
Deflections arise primarily in the whisker, which is much less stiff than the base bridge.
Computed deflections match observations to within 0.5\,mm.
Five different designs, including varying bridge lengths, $L \in \{$45\,mm$,$ 35\,mm$,$ 25\,mm$\}$, for a fixed bridge height $H=$ 2\,mm, and varying bridge heights, $H \in \{$2\,mm$,$ 4\,mm$,$ 6\,mm$\}$, for a fixed bridge length $L=$ 45\,mm, were evaluated (Video~3).
Figure \ref{fig:whisker_model}D and E depict the corresponding variation of $\Delta \lambda_{1}$ with respect to $\Delta F$ for these designs, with both simulated results (lines) and experimental results (light-colored bands).
The simulated results indicate an approximately linear relationship in all cases, revealing that the deformation of the base bridge, where the FBGs are located, remains within the small-deformation linear-elastic region, even when the whisker undergoes a large deflection, whose tip motion exceeds 30\,mm.
The experimental results showcase slight deviations.
This is attributed to several factors, including the non-linearity properties of the 3D-printing material and glue and minor slippage at the contact point.
However, the tendency of simulated and experimental results matches, verifying the usability of the established model in rapid estimation and design for whiskers with distinct geometric profiles.
We aimed at enhancing the sensitivity of the whisker by tuning different values for $H$ and $L$, subject to fabrication, base size, and durability constraints.
Through model simulations, sensitivity improved as $H$ decreased and as $L$ increased.
The selected design is $\{H,L\} = \{$2\,mm $,$ 45\, mm$\}$.
We did not further reduce $H$, as this poses challenges for the fabrication process when attaching the whisker to the base bridge, making it difficult to ensure consistency.
Our minimum load detection limit is $\approx$ 0.5\,mN, considering that the $\pm 3 \sigma$ noise level of the FBG wavelength fluctuation is approximately 2.7\,pm.
It is worth noting that this performance was achieved under the equivalent point-load scenario. Different loading conditions will correspond to different performance levels. In underwater applications, the whisker is less subject to thermal drift, which can improve performance.

\section{Experiments and Results}
This study is guided by two research questions:
(1) What hydrodynamic cues can the FBG whisker sensor measure under controlled relative-flow and foil-generated wake conditions?
(2) Can these cues support a constrained robotic decision-making task based on delayed wake information?
To address these questions, we performed three sets of experiments.
First, we characterized the whisker under controlled towing to quantify vortex-induced-vibration suppression, speed-dependent bending response, and angle-of-attack dependence.
Second, we tested the whisker's response to unsteady flows generated by a pitching NACA0012 foil, comparing a fixed pitching-foil condition with a co-translating finite-Strouhal-number condition that better approximates the delayed hydrodynamic cue left by a moving source.
Third, we integrated a single whisker on a mobile ROV and evaluated whether whisker signals alone could support binary branch selection in a delayed foil-generated trail.
Finally, we performed a field deployment in the ocean as a qualitative robustness check of wake detectability outside the controlled pool environment.
\subsection{Flow Sensing}
Before studying foil-generated wakes and robotic trail detection, we characterized the whisker under controlled towing to establish three basic sensing behaviors: suppression of VIV, response to relative flow speed, and dependence on angle of attack (AoA). These experiments define the operating behavior of the sensor under well-controlled relative flow.
\begin{figure*}[!t]
    \centering
    \includegraphics[width=\textwidth]{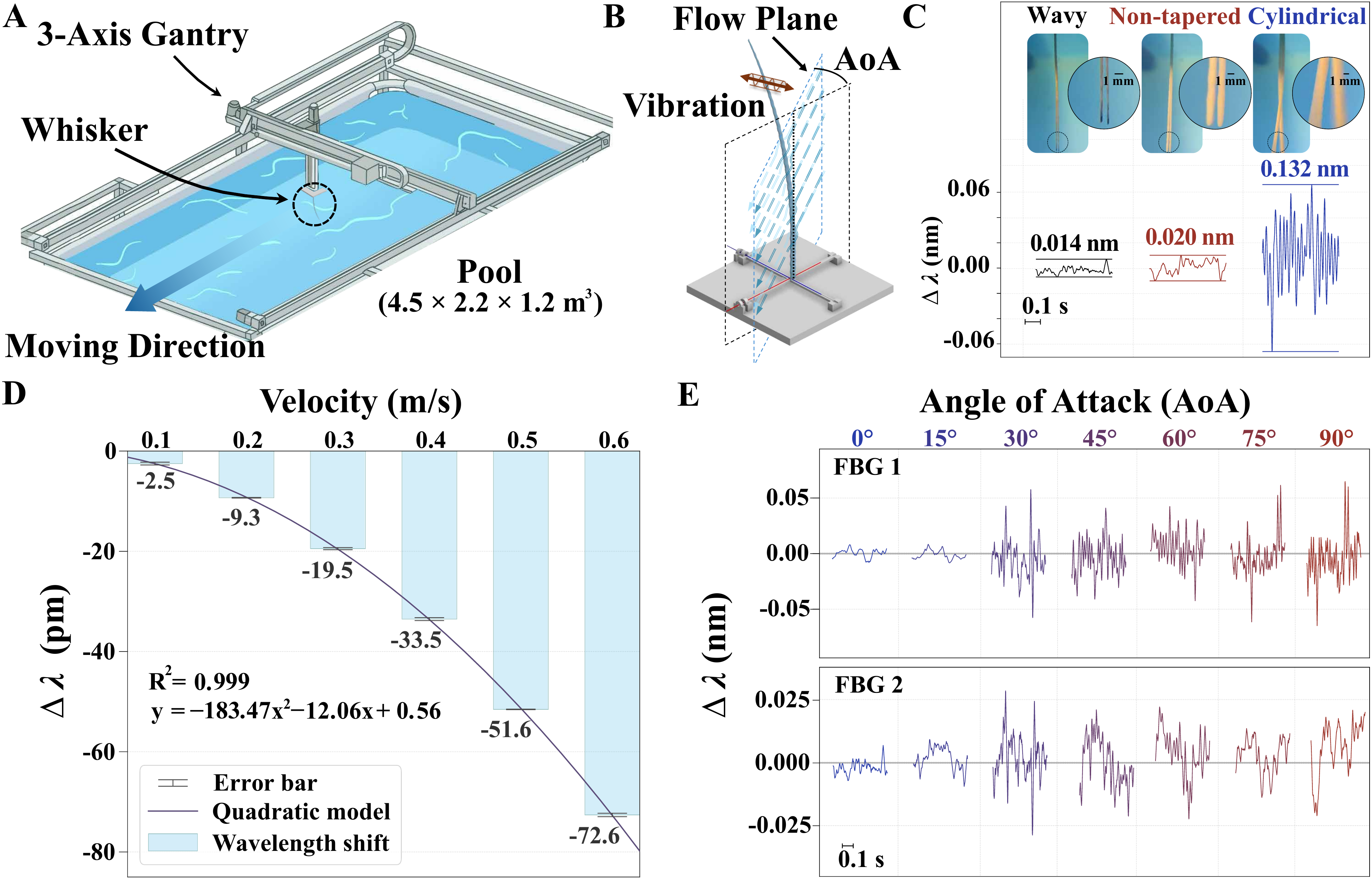}
    \caption{\textbf{Flow-sensing experiments.}
    (\textbf{A}) A three-axis gantry is used for controlled towing experiments in a pool. The whisker sensor was mounted on the $z$-axis end-effector and translated along the $x$-axis to generate prescribed relative flow.
    (\textbf{B}) Schematic of mean drag-induced bending and cross-flow oscillation caused by vortex-induced vibration (VIV). Angle of attack (AoA) is defined by the orientation of the whisker's narrow side relative to the incoming flow.
    (\textbf{C}) Three whisker geometries tested---wavy tapered, wavy non-tapered, and cylindrical---with representative mean-centered FBG~2 wavelength-shift traces at 0.6~m/s.
    (\textbf{D}) Mean baseline-referenced wavelength shift of the wavy tapered whisker versus towing speed. Error bars denote standard deviation across 5 replicate runs after the outlier-screening procedure described in Appendix~\ref{app:processing}.
    (\textbf{E}) Representative mean-centered two-channel whisker signals at different AoAs while towing at 0.6~m/s.}
    \label{fig:pool}
\end{figure*}
\subsubsection{Experimental setup}
All towing experiments were conducted in an outdoor, above-ground pool (4.5~m $\times$ 2.2~m $\times$ 1.2~m), as shown in Fig.~\ref{fig:pool}A. A three-axis gantry system (FUYU Technology Co., Ltd.) was mounted above the pool to provide repeatable, commanded motion. The whisker sensor was rigidly attached to the gantry z-axis end-effector and towed along the x-axis at prescribed constant speeds, thereby generating a controlled relative flow over the sensor. AoA is defined as the angle between the whisker's narrow side and the incoming flow (Fig.~\ref{fig:pool}B). In this towing configuration, FBG~1 mainly captures the mean bending response associated with quasi-steady hydrodynamic loading, whereas FBG~2 is more sensitive to cross-flow oscillations associated with VIV.
\subsubsection{Geometry comparison and VIV suppression}
We first compared three whisker geometries---wavy tapered, wavy non-tapered, and cylindrical---to evaluate their self-induced vibrations during towing. Representative trials are shown in Video~1. As shown in Fig.~\ref{fig:pool}C, the cylindrical whisker exhibits the largest cross-flow oscillations in the FBG~2 channel, whereas both wavy whiskers show substantially reduced fluctuations. The representative oscillation amplitude of the cylindrical whisker produces an FBG wavelength shift of $0.132$~nm, corresponding to a tip motion of about $5.4$~mm. By comparison, the wavy non-tapered and wavy tapered whiskers exhibit wavelength shifts of $0.020$~nm and $0.014$~nm, corresponding to tip motions of about $1.5$~mm and $0.9$~mm, respectively. These results demonstrate that the bio-inspired wavy geometry suppresses vortex-induced vibration relative to the cylindrical baseline.
\subsubsection{Response to relative flow speed}
We next quantified the relationship between whisker output and relative flow speed by towing the wavy tapered whisker at speeds from 0.1 to 0.6~m/s. As shown in Fig.~\ref{fig:pool}D, the mean baseline-referenced wavelength shift increased monotonically in magnitude with towing speed. The grouped mean response, expressed in pm, was fit with the quadratic calibration curve
\begin{equation}
\Delta\lambda_{\mathrm{pm}} = -183.47v^2 - 12.06v + 0.56,
\end{equation}
where $v$ is the commanded towing speed in m/s. This calibration establishes relative flow-speed sensitivity over the tested range; it is not intended as a universal calibration independent of whisker orientation, mounting, or flow environment.
\subsubsection{Effect of angle of attack}
We evaluated the effect of AoA by rotating the wavy tapered whisker from $0^\circ$ to $90^\circ$ in $15^\circ$ increments while towing at 0.6~m/s. Representative trials are provided in Video~2. As shown in Fig.~\ref{fig:pool}E, both FBG channels exhibit only weak fluctuations at $0^\circ$ and $15^\circ$, whereas the oscillations become much more pronounced from $30^\circ$ onward. This trend is consistent with the increase in projected cross-flow area as the whisker rotates relative to the incoming flow. The result shows that whisker response depends strongly on orientation and that AoA should be taken into account when comparing trials or interpreting robot-mounted measurements.
\begin{figure*}[!t]
    \centering
    \includegraphics[width=\textwidth]{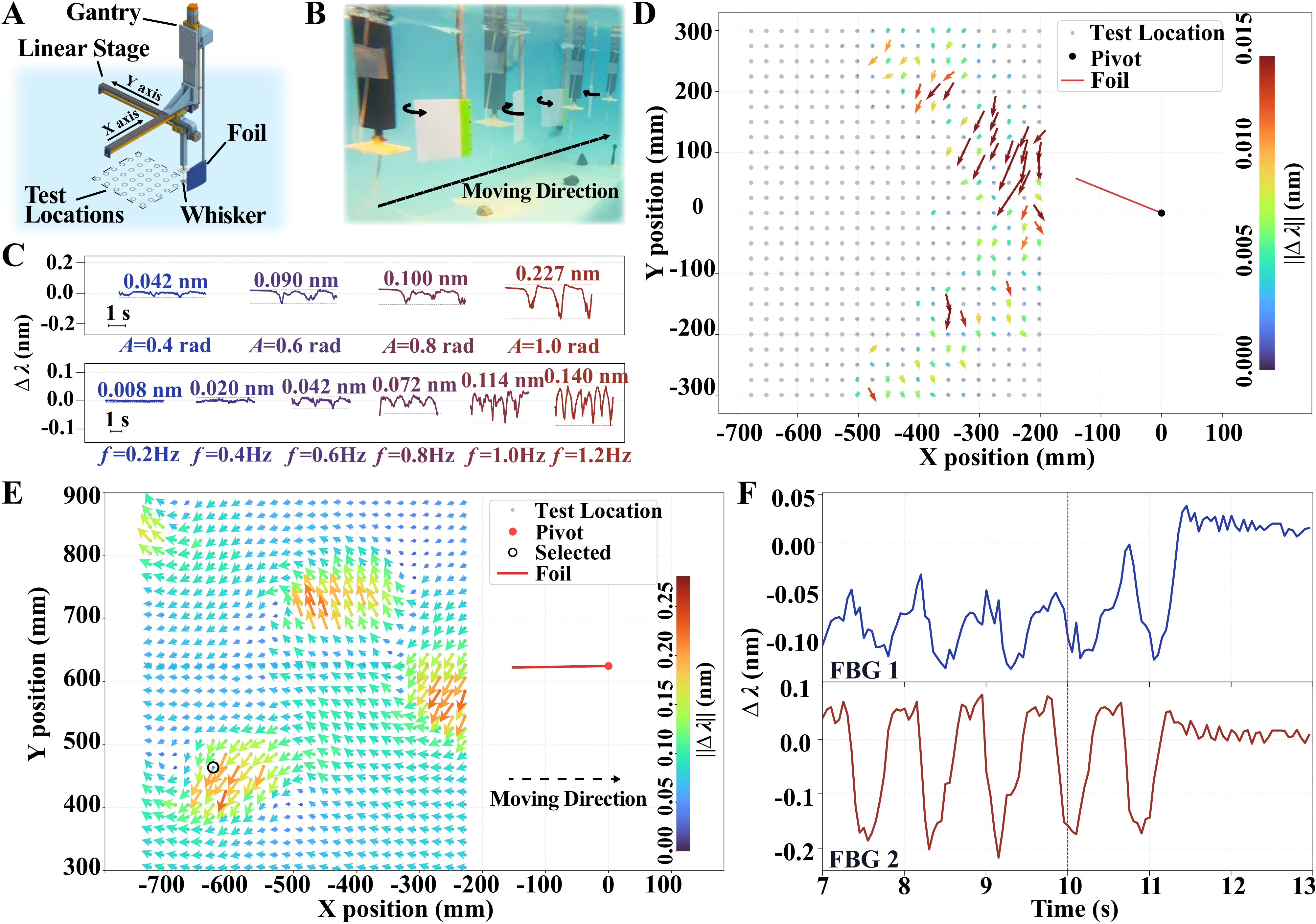}
    \caption{\textbf{Wake sensing with a pitching foil.}
    (\textbf{A}) Setup for pitching foil wake sensing. A positioning gantry mounted to the $z$-axis of the main gantry adjusted the whisker position relative to a NACA0012 foil. Whisker measurements were collected over a prescribed scanning grid, with one whisker sensor placement corresponding to each grid location.
    (\textbf{B}) Co-translating foil-whisker configuration used to approximate sensing of the wake left by a moving source.
    (\textbf{C}) Whisker response to foil pitching at different amplitudes and frequencies, measured at $x=-100$\,mm and $y=100$\,mm relative to the foil tip at its initial position. Displayed traces are mean-centered for visual comparison.
    (\textbf{D}) Whisker response map for a pitching foil and a stationary whisker in quiescent water. Each arrow represents
    the sensor signal from a single test with the foil at the corresponding angle. Hence, the vectors are synchronized in terms of phase but are not simultaneous measurements.
    (\textbf{E}) Reconstructed whisker-response map in the co-translating finite-\(\mathrm{St}\) regime. The map was assembled from repeated single-location trials; the full displayed domain was completed from two acquisition batches using the reflection procedure described in Appendix~\ref{app:processing}, which assumes approximate symmetry about the trajectory centerline.
    (\textbf{F}) Representative two-channel FBG signals at the location marked, $\circ$, in panel E. In (D) and (E), arrows show the two-channel response vector formed from FBG 1 and FBG 2, and color encodes the vector magnitude. These arrows represent reconstructed FBG response vectors at sampled or mirrored relative locations, not simultaneous velocity vectors or direct measurements of vorticity.}
    \label{fig:vortex_exp}
\end{figure*}
\subsection{Wake-Generator Regimes and Wake Sensing}
The towing experiments above establish the whisker's response to controlled relative flow. We next examine foil-generated unsteady flow, which is the hydrodynamic cue most relevant to wake sensing and to the later robot branch-selection task. Because the disturbance field depends strongly on source kinematics, we distinguish between a pitching foil held fixed in quiescent water and a pitching foil that translates relative to the water.
Many artificial whisker studies use an upstream cylinder, a pitching foil, or a fishtail as a convenient disturbance source. Such tests establish whether a sensor responds to unsteady flow, but they do not necessarily reproduce the downstream-convected disturbance left by a translating source. A useful nondimensional parameter for translating oscillatory sources is the Strouhal number,
\begin{equation}
    \mathit{St} = \frac{f A_{\mathrm{TE}}}{U},
    \label{eq:strouhal}
\end{equation}
where $f$ is the pitching frequency, $A_{\mathrm{TE}}$ is the peak-to-peak trailing-edge excursion, and $U$ is the translation speed of the source relative to the surrounding water.
For a rigid symmetric foil pitching at a fixed location in quiescent
water, corresponding to $St \to \infty$, prior fluid-mechanics studies report a laterally meandering jet rather than a stable downstream-convected reverse
von K\'arm\'an wake~\cite{shinde2013jet,shinde2018physics}. Conversely, if the goal is to approximate delayed sensing of the hydrodynamic disturbance left by a moving source, a translating finite-$St$ regime is a closer experimental analogue. Efficient oscillatory propulsion is commonly associated with Strouhal numbers of order $0.2$--$0.4$, with values near $0.3$ often reported~\cite{triantafyllou1991wake,taylor2003flying}. We compare both conditions below and use that comparison to motivate the operating point for the robot branch-selection experiments.
\subsubsection{Experimental setup}
As shown in Fig.~\ref{fig:vortex_exp}A, a small positioning gantry was attached to the $z$-axis of the main gantry so that the whisker position could be adjusted relative to a NACA0012 foil. The whisker response was recorded with the two FBG channels, and spatial maps were assembled by repeating trials over a grid of relative foil positions. In the fixed-pitching case, there is no mean translation. In the co-translating case, shown in Fig.~\ref{fig:vortex_exp}B, the foil and whisker move together at a prescribed Strouhal number, defined from the foil kinematics and translation speed in Section~C of Appendix~B.
\subsubsection{Response to imposed oscillatory forcing}
Before comparing wake regimes, we verified that the whisker responds measurably to foil-generated unsteady flow by varying the pitching angle, $A$,
and frequency, $f$, while holding the whisker at a distance $x=-100$\,mm and $y=100$\,mm relative to the foil tip at its initial position.
The response amplitudes are shown in Fig.~\ref{fig:vortex_exp}C as shifts in FBG wavelengths.
\subsubsection{Fixed pitching foil in quiescent water}
We first mapped the whisker response around a pitching foil held at a fixed location while the whisker was also stationary. For this map, the foil was pitched at $f=0.6$\,Hz with angular amplitude $A=0.4$\,rad for eight cycles, while the relative foil position was varied over a spatial grid. The resulting response map, shown in Fig.~\ref{fig:vortex_exp}D and animated in Video~6, shows stronger signals at lateral offsets from the foil centerline and weaker responses directly downstream. In the absence of mean advection, this configuration is not expected to generate the downstream-convected trail most relevant to delayed branch selection behind a moving source. This case is therefore useful as a controlled unsteady-flow benchmark, but not by itself a direct proxy for the hydrodynamic trail left by a moving target.
\subsubsection{Translating foil trail analogue}
To better approximate delayed sensing behind a moving source, we next translated the foil and whisker together in the same direction while the foil continued to pitch. In this configuration, both the foil and whisker moved at 0.46\,m/s, while the foil pitched with angular amplitude $A = 0.4$\,rad and frequency $f=1.2$\,Hz. Using the trailing-edge excursion defined in Section~C of Appendix~B, this operating point corresponds to $St=0.35$. This value lies within the commonly cited range associated with oscillatory propulsion and thrust-producing wakes~\cite{triantafyllou1991wake,taylor2003flying}, making this condition a closer experimental analogue for delayed sensing of a moving-source trail than the fixed-pitching case.
The reconstructed whisker-response map in Fig.~\ref{fig:vortex_exp}E, with an animated version in Video~7, shows clustered regions of elevated whisker response and alternating local vector directions. Because the map was assembled from repeated single-location trials and, for the full moving-source domain, from two acquisition batches using the reflection procedure described in Appendix~\ref{app:processing}, it should be interpreted as a response-field reconstruction rather than a simultaneous planar flow-field measurement. These patterns are consistent with repeated encounters with convected unsteady structures, but the vector plots are whisker-response maps rather than direct velocity or vorticity measurements. A representative signal measured at the marked location, shown in Fig.~\ref{fig:vortex_exp}F, exhibits oscillatory signatures in both FBG channels. Compared with the fixed-pitching case, this co-translating condition produces a more organized downstream response and is therefore the regime used in the robot branch-selection experiments that follow.
\subsubsection{Connection to robot branch selection}
The comparison above motivates the wake-generator settings used in the robotic experiments. Because the downstream cue in the robot task is generated by a moving foil, the co-translating finite-$St$ condition is the more appropriate pool analogue. We therefore use this regime in the branch-selection experiments below, where the foil first generates a delayed hydrodynamic cue and the robot uses whisker measurements to choose between candidate branches after the source has moved on.
\subsection{Whisker-Guided Branch Selection}
\begin{figure*}[!t]
    \centering
    \includegraphics[width=\textwidth]{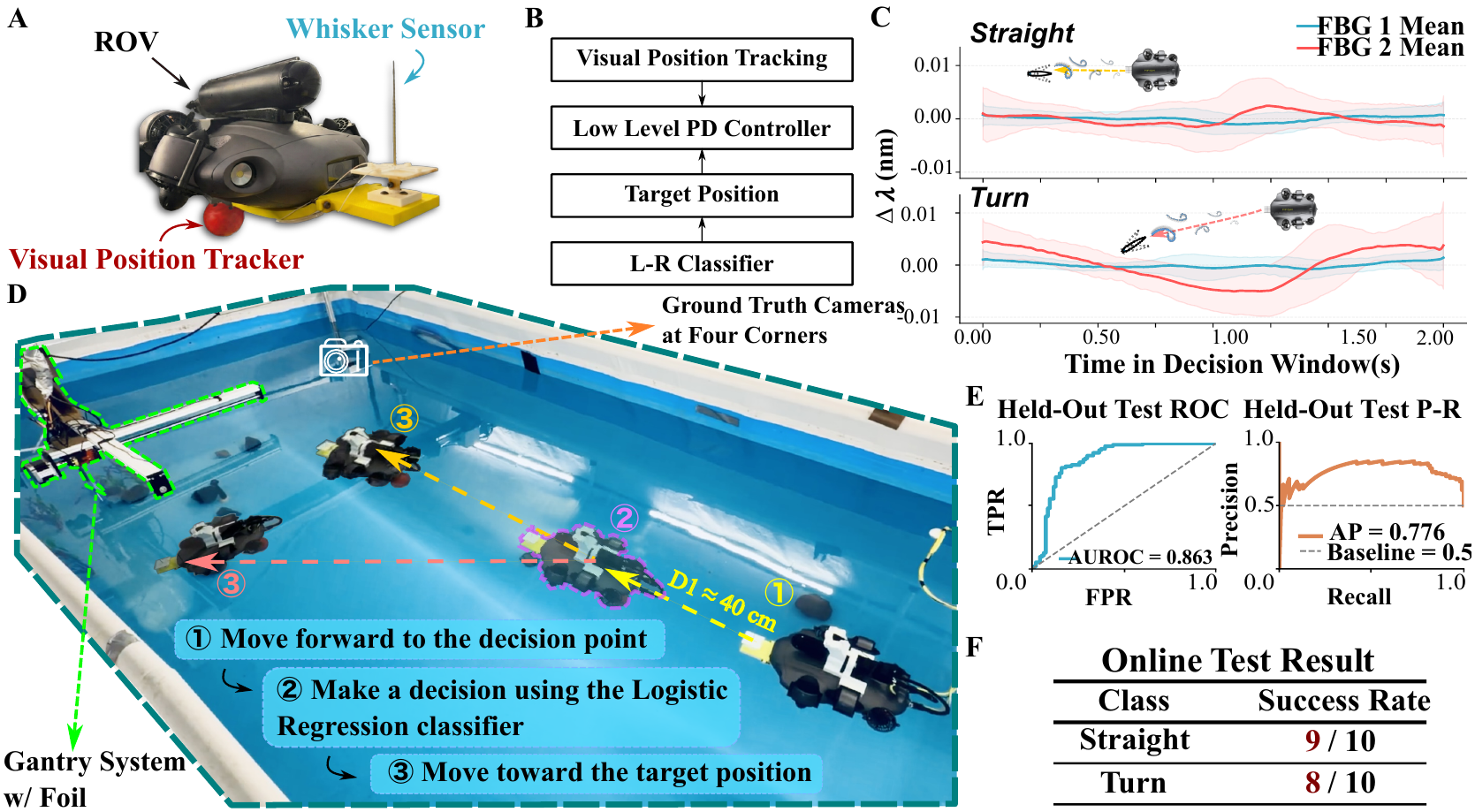}
    \caption{\textbf{Whisker-guided branch selection using a delayed foil-generated trail.}
    (\textbf{A}) Commercial ROV with a front-mounted whisker; the visual position tracker was used for vehicle localization and evaluation, but not as an input to the whisker classifier.
    (\textbf{B}) Control and decision pipeline: localization supported low-level vehicle motion, while a logistic-regression classifier selected the target branch from whisker-derived features.
    (\textbf{C}) Mean two-channel FBG signals within the 2\,s decision window for straight and turning runs, with shaded $\pm$ one standard deviation envelopes computed over held-out test windows.
    (\textbf{D}) Pool experiment showing gantry-mounted foil, fixed start point, decision region, and candidate branches. The ROV first moved forward to the decision point, classified the branch using the logistic-regression classifier, and then moved toward the selected target position. The start-to-decision distance is denoted by $D_1 \approx 0.40~\mathrm{m}$.
    (\textbf{E}) Receiver-operating-characteristic and precision--recall curves for the offline classifier on held-out test windows.
    (\textbf{F}) Online branch-selection success rates for straight and turning trials.}
    \label{fig:tracking}
\end{figure*}

We next tested whether whisker signals alone could allow a robot to distinguish between two candidate paths in a delayed foil-generated trail. Using the wake-generator settings motivated above, the foil first traversed the pool to lay a repeatable hydrodynamic trail. The robot was then released after a short delay from a fixed start point and advanced toward a predefined decision region, where it had to choose between two candidate branches---continuing straight or executing a turn---using whisker input alone (Fig.~\ref{fig:tracking}D). The start point was approximately 0.40\,m behind the foil's initial position along the nominal path; as a practical benchmark for delay in this geometry, about 5\,s elapsed between foil-motion onset and the robot reaching that position. This branch-point formulation was chosen as a controlled first step toward delayed wake-guided navigation, rather than as a demonstration of continuous trail centering or biological wake tracking.
A single whisker sensor was mounted on the front of a commercial underwater robot (QYSea V6 Expert; Fig.~\ref{fig:tracking}A). A red spherical marker beneath the vehicle was tracked optically to provide vehicle position feedback for the PD controller during data collection and evaluation. These visual-tracking signals were used for trajectory execution and for identifying the robot's position during offline window selection, but they were not provided as input features to the branch classifier.
For labeled data collection, the foil was initially positioned approximately 0.4\,m ahead of the robot and was commanded to follow either a straight path or a turning path. After the foil began moving and generated a hydrodynamic trail, the robot was released after a short delay. The position-feedback controller first drove the robot straight forward for 0.5\,m, bringing it to the region where the foil trajectory separated into the two candidate branches. The robot was then commanded to follow the same branch taken by the foil, either continuing straight or turning, while the two FBG channels were recorded throughout the run.
\begin{table*}[!t]
\footnotesize
\centering
\caption{\textbf{Representative underwater whisker sensors.} Flow speed lists nonzero validation conditions; wake distance is reproduced as reported and, when available, normalized by $d_w$. APPENDIX~A SUMMARIZES THE COMPARISON CONVENTIONS.}
\label{tab:whisker-comparison}
\resizebox{\textwidth}{!}{
\begin{tabular}{l l c c c c l}
\toprule
\textbf{Reference} & \textbf{Principle} & \textbf{$L\times d_1\times d_2$ (mm)} & \textbf{Flow speed (m/s)} & \textbf{Wake source} & \textbf{Wake distance} & \textbf{System demonstration} \\
\midrule
Beem et al.\ \cite{beem2012calibration, beem2015wake} & Resistive & 275 $\times$ 35.7 $\times$ 14.4 & 0.2--0.8 & Cylinder & $160\,d_w$ & Benchtop wake sensing \\
Liu et al.\ \cite{liu2023artificial} & Resistive & 35 $\times$ 2.14 $\times$ 1.06 & 0.048--0.25 & Cylinder & 250 mm ($\sim 235\,d_w$) & Benchtop wake sensing \\
Zheng et al.\ \cite{zheng20223d} & Resistive & 150 $\times$ 11.6 $\times$ 5.0 & 0.1 & Cylinder & 80 mm ($\sim 10\,d_w$) & Benchtop wake sensing \\
Guo et al.\ \cite{guo2024piezoelectric} & Piezoelectric & 210 $\times$ 15 $\times$ 7.5 & 0.25--0.5 & Cylinder & 240 mm ($\sim 16\,d_w$) & Benchtop wake sensing \\
Wang et al.\ \cite{wang2022underwater} & TENG & 150 $\times$ N/A $\times$ N/A & N/A & Fishtail & 750 mm & Benchtop wake sensing \\
Dai et al.\ \cite{dai2024biomimetic} & Magnetic & 10.92 $\times$ N/A $\times$ N/A & max $0.6^{\dagger}$ & Fishtail & 150 mm ($\sim 95\,d_w$) & Array sensing \\
Elshalakani et al.\ \cite{elshalakani2020deep} & Optical & 100 $\times$ 0.75 $\times$ 0.75 & 0.3 &  Cylinder & 875 mm & Wake-source localization \\
Glick et al.\ \cite{glick2024tracking} & FBG & 80 $\times$ 0.25 $\times$ 0.25 & 0.1--0.25 & Cylinder & 750 mm & Benchtop wake sensing \\
Wang et al.\ \cite{wang2024bioinspired} & FBG & 130 $\times$ 7.14 $\times$ 2.88 & max $0.36^{\dagger}$ & Fishtail & 300 mm ($\sim 95\,d_w$) & Benchtop wake sensing \\
Xu et al.\ \cite{xu2024deep, liu2025deep} & TENG & 200 $\times$ 12.4 $\times$ 5.0 & max $0.6^{\dagger}$ &  Cylinder & 40 mm & ROV state estimation \\
Our design & FBG & 100 $\times$ 3.15 $\times$ 1.068 & 0.1--0.6 & Moving foil & 576.44~mm$^{\ddagger}$ ($\sim 484\,d_w$) & Online branch selection \\
\bottomrule
\end{tabular}
}
\vspace{2pt}
\begin{minipage}{0.99\textwidth}
\footnotesize
$^{\dagger}$ The smallest positive steady-flow speed was not unambiguously recoverable from the source paper; only the maximum reported or tested positive speed is therefore listed.\par
$^{\ddagger}$ For our design, 576.44~mm is the maximum co-translating source--sensor separation sampled within the available experimental workspace, not an independently measured detection limit.
\end{minipage}
\end{table*}
Training samples were extracted from the decision region immediately before the robot executed the branch maneuver. Specifically, for each trajectory we selected timestamps at which the robot's forward displacement was between 0.45\,m and 0.50\,m from its start position. For each selected timestamp, we extracted a 2\,s window of the two-channel FBG signal ending at that timestamp. This procedure produced multiple labeled windows from each trajectory, with labels assigned according to the foil path, straight or turning. Demonstrations were partitioned at the trajectory level before sliding-window extraction, yielding 50 training demonstrations, 25 per class, 10 validation demonstrations, and 20 held-out test demonstrations. We trained a logistic-regression classifier on engineered features extracted from both FBG channels; preprocessing and feature construction are described in Appendix~\ref{app:model-classifier}.
Mean signals within the decision window are shown in Fig.~\ref{fig:tracking}C. Relative to straight runs, turning runs tend to exhibit a stronger oscillatory buildup within the decision window, especially toward its end. On held-out test windows, the classifier achieved an Area Under the Receiver Operating Characteristic Curve (AUROC) of 0.863, an average precision (AP) of 0.776, and 77.4\% accuracy; the corresponding training and validation accuracies were 92.2\% and 73.7\%, respectively (Fig.~\ref{fig:tracking}E). These window-level results indicate that whisker signals contain a branch-discriminative hydrodynamic cue, while also suggesting some overfitting consistent with the modest dataset size.
The online experiment provides a more task-relevant run-level evaluation. Once the robot entered the decision region, the classifier produced a single prediction from the most recent 2~s whisker history, and the robot committed to the corresponding branch for the remainder of the run. Across 20 run-level trials, the robot selected the correct branch and reached the corresponding target region in 17 cases (85.0\%), as shown in Fig.~\ref{fig:tracking}F; representative runs are shown in Video~9. Thus, even a single whisker provided a hydrodynamic cue sufficient for delayed binary branch selection several seconds after wake generation.
This result should be interpreted within the scope of the present task. The experiments were conducted in a controlled pool, with a single whisker, largely planar motion, and a binary branch decision rather than continuous trail centering. We therefore view this experiment as an initial systems demonstration of delayed, whisker-guided branch selection in a foil-generated trail, not yet as continuous trail following or a robotic analogue of biological wake tracking.

\subsection{Field Deployment}
We also conducted an open-water deployment at a coastal marine field site to evaluate whether the whisker signals remained detectable outside the controlled pool environment. The goal of this test was not closed-loop trail tracking, but a field robustness check: can a whisker-equipped ROV detect vortical disturbances generated by a nearby swimmer in coastal water, where ambient wave, current, and vehicle-induced motions are present? In the representative trial summarized in Fig.~\ref{fig:overall}D and Video~10, a human diver swam near the vehicle and generated a fin wake while the robot recorded the two FBG channels. The diver and vehicle moved at speeds on the order of 0.2~m/s, with a separation of approximately 0.6~m; the diver's fins oscillated at approximately 1.35~Hz with an angular amplitude of about $30^\circ$. The whisker signal showed a distinct two-channel event after the fin passage, with an estimated delay of $\delta t \approx 1.2$~s between wake generation and detection. This result indicates that the FBG whisker can detect externally generated vortical disturbances in a coastal setting, although the trial should be interpreted as a qualitative deployment demonstration rather than a calibrated wake-range or closed-loop tracking measurement.

\section{Discussion and Limitations}
The main contribution of this work is not only a new whisker transducer, but an experimental progression from biological principles to robot behavior. Harbor seals do not use their vibrissae merely to detect generic unsteady flow; they use them to interpret delayed hydrodynamic trails left by moving prey. Reproducing that task in robotics, therefore, requires more than a sensitive whisker. It requires reducing self-induced vibration, establishing basic flow-sensing capabilities, selecting a wake-generation regime relevant to delayed trail following, and showing that the resulting signals can initiate action on a mobile platform. The present study addresses that sequence within a single robot-compatible FBG whisker system.
More broadly, artificial whiskers have established important sensing capabilities, including steady-flow estimation, directional response, wake detection, wake-source localization, and vehicle-state estimation~\cite{beem2015wake,wang2022underwater,elshalakani2020deep,glick2024tracking,wang2024bioinspired,xu2024deep,liu2025deep}. Because nonzero steady-flow speed coverage, wake distance, sensitivity, response time, and SNR are defined and reported differently across studies, we treat Table~\ref{tab:whisker-comparison} as a contextual comparison rather than as a rank ordering. In particular, a reported $0$~m/s condition is treated as a no-flow baseline rather than as evidence of a measurable lower flow-speed limit; the corresponding extraction rules and comparison conventions are summarized in Appendix~A. Our advance is to connect these ingredients in a robot-compatible FBG implementation and to show that whisker signals can support an online behavioral choice on a mobile underwater robot after the wake source has moved on.
Several limitations remain. The wake source is a simplified foil-like generator rather than a freely swimming animal; the robot solves a binary branch-selection problem rather than continuous trail centering; the experiments were conducted in a controlled pool; and the present system uses a single whisker in an approximately planar setting. Together with the modest number of online trials, these constraints argue for careful scoping of the claim: we demonstrate whisker-guided delayed branch selection in a controlled task, not robust free-form wake tracking. Future work should extend this framework to multi-whisker arrays, more realistic source kinematics, background currents, and three-dimensional vehicle motion. These directions will be important for separating self-induced flow from external hydrodynamic cues and for approaching the robustness of biological trail following. The passive optical readout and multiplexing capability of FBGs make that scaling direction especially attractive~\cite{massari2019fbg,Massari2022Ruffini,delatorre2021underwaterFBG}.

\appendices
\section{Geometry, Robustness, and Cross-Study Conventions}
\label{app:methods}

\subsection{Whisker Geometry}
The undulated cross-section follows the parameterization of Hanke \emph{et al.}~\cite{hanke2010viv}, with a quadratic taper applied from base to tip. For each semi-axis $s_0\in\{a,b,k,l\}$, the value at cross-section index $n$ is
\begin{equation}
 s(n)=s_0\left[1-(n t_0)^2\right],
\end{equation}
where $t_0=0.029$. The dimensions used in the CAD model and fabrication workflow are listed in Table~\ref{tab:whisker-morphology-params}.

\begin{table}[!t]
\caption{Whisker geometry used in the experiments. Peak and trough dimensions are elliptical semi-axes at the base.}
\label{tab:whisker-morphology-params}
\centering
\footnotesize
\begin{tabular}{lcc}
\toprule
Parameter & Symbol & Value \\
\midrule
Peak semi-axes & $a\times b$ & $1.575\times0.534$ mm \\
Trough semi-axes & $k\times l$ & $1.248\times0.657$ mm \\
Inclination angles & $\alpha,\beta$ & $15.27^\circ,17.6^\circ$ \\
Undulation period & $\Lambda$ & $3.5$ mm \\
Taper offset & $t_0$ & $0.029$ \\
Total length & $L_t$ & $\sim100$ mm \\
\bottomrule
\end{tabular}
\end{table}

\subsection{Durability and Large-Deflection Recovery}
\begin{figure}[!t]
\centering
\includegraphics[width=\columnwidth]{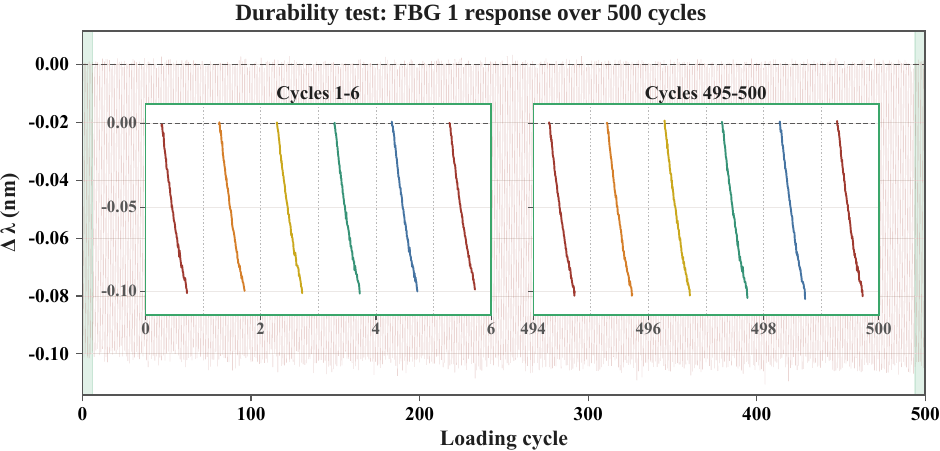}
\caption{Cyclic-loading durability of the selected whisker design. FBG~1 wavelength shifts recorded over 500 consecutive indentation cycles. Insets compare representative early and late cycles, showing no obvious loss of mechanical-to-optical response over the tested sequence.}
\label{fig:durability-cycles}
\end{figure}

The selected design was subjected to 500 repeated indentation cycles on a linear stage while FBG~1 was recorded; the early and late-cycle waveforms show no obvious loss of mechanical-to-optical response over the tested sequence. The corresponding FBG~1 traces over all 500 loading cycles are shown in Fig.~\ref{fig:durability-cycles}. In three underwater release trials, the shaft was manually deflected beyond $90^\circ$ at progressively more distal contact locations while both FBG channels were sampled at approximately 2~kHz. Force onset was defined from the differential signal $\lambda_1-\lambda_2$ as the first post-baseline sample exceeding ten baseline standard deviations. The peak excursion was the largest absolute FBG~1 deviation in the next 8~s. Recovery to the $\pm5\%$ peak band required that the signal remain within that band for 1~s. Peak excursions were 1643, 673, and 512~pm, and the corresponding $\pm5\%$ amplitude-recovery times were 0.87, 0.77, and 0.50~s. The recovery times annotated in Video~4 refer instead to a separate 5~s sliding-window F-test criterion for variance recovery; for the same three trials, the corresponding F-test recovery times were 7.86, 4.84, and 3.80~s, respectively. The residual mean offsets at those times were 0.4, 2.8, and 1.4~pm, comparable to the approximately 1.1~pm baseline standard deviation (Videos~4 and~5).

\subsection{Comparison Conventions}
The literature table reports the smallest and largest positive steady-flow speeds that could be recovered from each source; a 0~m/s trace is treated as a static baseline rather than a demonstrated lower measurement limit. Wake distance is reported in the source paper's absolute units and, when available, in its own whisker-diameter convention. Sensitivity, response time, and signal-to-noise ratio are protocol dependent and are not treated as directly rankable across modalities. For this work, response time is the interval from the first 240~Hz video frame showing physical contact to the first clear FBG departure from baseline (29.74~ms), and steady-flow SNR is $20\log_{10}(|\mu_{\mathrm{steady}}-\mu_{\mathrm{static}}|/\sigma_{\mathrm{static}})$.

\section{Signal Processing and Wake-Map Reconstruction}
\label{app:processing}

\subsection{Towing Experiments}
For the geometry comparison, FBG~2 traces were linearly interpolated to 1~kHz, baseline corrected, and smoothed with a third-order Savitzky--Golay filter using a 31-sample window. A common fixed-duration window inside the stable-velocity portion of each run was used for all geometries; no geometry-specific peak search was performed. Traces were centered by subtracting their midrange for display, which does not change peak-to-peak amplitude.

For the speed sweep, the FBG~1 baseline for each run was the mean over $t=0$--4~s, and the response was the mean of a predefined constant-speed interval minus that baseline. Five runs were acquired at each commanded speed from 0.1 to 0.6~m/s. Run-level responses were screened within each speed using a median-absolute-deviation modified-$z$ threshold of $|z|\leq4$; the plot reports the mean and standard deviation of retained runs. The second-order curve is a visualization fit to grouped means. For angle of attack, both channels were resampled to 1~kHz and processed with the same 31-sample Savitzky--Golay filter; identical absolute-time windows in the stable-motion region were used for all seven angles.

\subsection{Wake-Response Maps}
Each point in the displayed maps was obtained from a repeated single-location run and therefore is not a simultaneous planar flow measurement. For the fixed-foil map, channels were baseline corrected using the first 1000 resampled samples and low-pass filtered at 10~Hz with a zero-phase fourth-order Butterworth filter. The two channels form the displayed response vector and their Euclidean norm sets the magnitude. For the moving-source map, runs were aligned to commanded foil-motion onset and no low-pass filter was applied so that transient peaks were retained. Two acquisition batches with opposite initial foil phase were combined because the gantry could not sample both sides in one sequence. The second batch was reflected about the trajectory centerline, $y'=2y_c-y$ with $y_c=625$~mm, and the transverse FBG component was sign-inverted. This reconstruction assumes approximate centerline symmetry. Samples were placed on a regular 16~mm by 4~mm grid and resampled to a common 20~Hz animation time base.

\subsection{Wake Kinematics}
For the co-translating case, foil and whisker moved at $U=0.46$~m/s while the foil pitched at $f=1.2$~Hz with angular amplitude $A_\theta=0.4$~rad. We define $\mathrm{St}=fA_{\mathrm{TE}}/U$, where $A_{\mathrm{TE}}$ is trailing-edge excursion; the operating point used for robot experiments gives $\mathrm{St}=0.35$.

\section{Model Implementation and Branch Classifier}
\label{app:model-classifier}

\subsection{Discretized Elastic Model}
\begin{figure}[!t]
\centering
\includegraphics[width=\columnwidth]{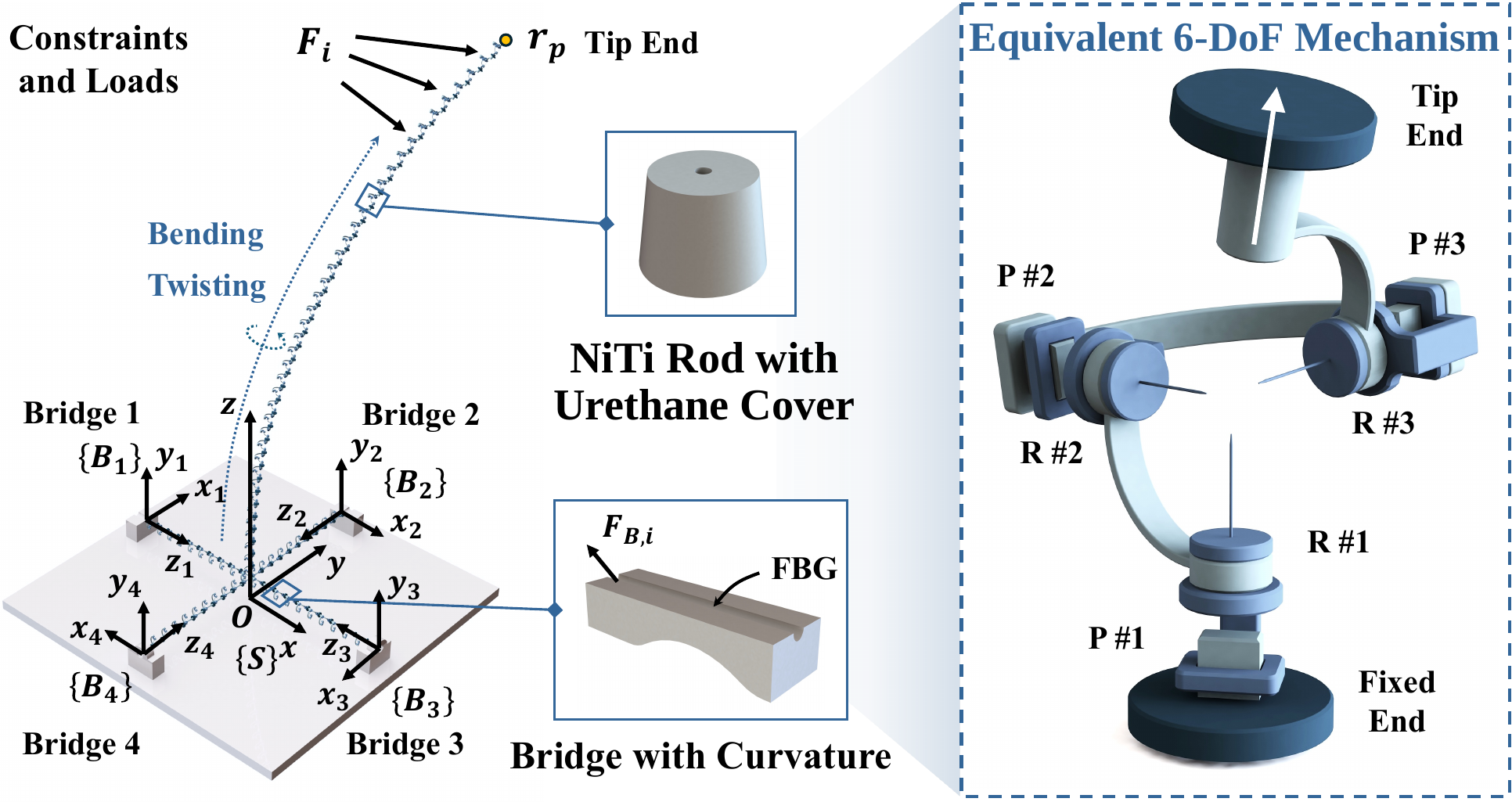}
\caption{Discretization-based analytical model. The whisker and the four base-bridge branches are divided into elastic segments. Each segment is represented by an equivalent six-DoF serial mechanism comprising three prismatic and three revolute joints, with stiffness assigned from local geometry and material properties. Coordinate frames are defined for the whisker and individual bridge branches, yielding a coupled parallel--serial hybrid mechanism subject to external loads $\vect{F}_{i}$ and interaction loads $\vect{F}_{B,i}$.}
\label{fig:analytical_model_framework}
\end{figure}

The whisker and each of four bridge branches were divided into short equal-length segments represented by equivalent six-degree-of-freedom elastic mechanisms. Figure~\ref{fig:analytical_model_framework} illustrates the corresponding coordinate frames, external loads $\vect{F}_{i}$, and whisker--bridge interaction loads $\vect{F}_{B,i}$, together with the equivalent six-DoF mechanism used to represent each segment. Each equivalent segment comprises three prismatic and three revolute joints, and the discretized whisker and four bridge branches form a coupled parallel--serial hybrid mechanism. Segment stiffness was computed from local geometry and material properties. For the whisker, $E_{\mathrm{NiTi}}\approx60$~GPa and $\nu_{\mathrm{NiTi}}\approx0.33$ were used; the urethane contribution was neglected because its estimated stiffness contribution was below 0.15\%. Geometric compatibility and elastic equilibrium of the whisker and bridge were solved together using analytic Jacobians and Newton--Raphson iteration. This formulation captures large whisker deflection while retaining the small-strain bridge regime used for FBG transduction.

\subsection{Classifier Data and Preprocessing}
The binary classifier used only FBG~1 and FBG~2; robot pose, velocity, yaw, and visual tracking were excluded. A trailing 2.0~s window was extracted at each manifest timestamp. Splits were made at the trajectory level before window extraction to prevent leakage, producing 463 training, 114 validation, and 186 held-out test windows. Each channel was linearly resampled to 500~Hz, padded at the record boundary with the earliest available value when needed, detrended, mean centered, and zero-phase low-pass filtered at 25~Hz with a sixth-order Butterworth filter.

Each window produced 42 features: 21 per channel. These included four positive-lag autocorrelation descriptors, five Welch-PSD band powers plus dominant frequency below 25~Hz, and segment-wise RMS, means, and successive RMS ratios over four equal temporal segments. Features were standardized using training data only. The classifier was logistic regression with an $\ell_1$ penalty, \texttt{liblinear} solver, $C=0.1$, and a fixed 0.5 decision threshold; no class reweighting or post-hoc threshold tuning was used. The frozen model obtained 77.4\% accuracy, 0.863 AUROC, and 0.776 average precision on held-out windows. Sixteen of 42 coefficients were nonzero. The online evaluation used one first-decision window per run and achieved 17 correct decisions in 20 runs; this run-level result is the behaviorally relevant evaluation because adjacent offline windows overlap.

\FloatBarrier

\FloatBarrier
\bibliographystyle{IEEEtran}
\bibliography{references}
\end{document}